# A Deep Learning Approach for the solution of Probability Density Evolution of Stochastic Systems

Seid H. Pourtakdoust[1*], Amir H. Khodabakhsh[2]

[1] Center of Excellence in Aerospace Systems, Sharif University of Technology
Tehran 14588-89694, Iran
ORCID: 0000-0001-5717-6240

[2] Department of Aerospace Engineering, Sharif University of Technology, Tehran, Iran.
ORCID: 0000-0002-0457-8673

pourtak@sharif.edu, khodabakhsh@ae.sharif.edu

## Abstract

Derivation of the probability density evolution provides invaluable insight into the behavior of many stochastic systems and their performance. However, for most real-time applications, numerical determination of the probability density evolution is a formidable task. The latter is due to the required temporal and spatial discretization schemes that render most computational solutions prohibitive and impractical. In this respect, the development of an efficient computational surrogate model is of paramount importance. Recent studies on the physics-constrained networks show that a suitable surrogate can be achieved by encoding the physical insight into a deep neural network. To this aim, the present work introduces DeepPDEM which utilizes the concept of physics-informed networks to solve the evolution of the probability density via proposing a deep learning method. DeepPDEM learns the General Density Evolution Equation (GDEE) of stochastic structures. This approach paves the way for a mesh-free learning method that can solve the density evolution problem without prior simulation data. Moreover, it can also serve as an efficient surrogate for the solution at any other spatiotemporal points within optimization schemes or real-time applications. To demonstrate the potential applicability of the proposed framework, two network architectures with different activation functions as well as two optimizers are investigated. Numerical implementation on three different problems verifies the accuracy and efficacy of the proposed method.

---

[*] Corresponding Author: Tel: (+98) 21 6616 4610, Email: pourtak@sharif.edu

# 1 Introduction

In reality, the performance of many complex structures may be time-varying and not predictable due to inherent uncertainties associated with external loads, dimensions, and material properties. These uncertainties could cause undesirable performance characteristics or even lead to catastrophic failures. In this respect, propagation of the uncertainties via probability-based analysis methods can help one to predict and enhance the dynamic behavior and performance of stochastic structures. Given the time-varying nature of probabilistic models of complex structures, one can estimate the probability of failure, time-varying reliability, or even determine the sensitivity of these quantities with respect to uncertainties or control inputs. This information is not only useful to designers for a safer design but also can be used for processing optimal control policies of stochastic structures, establishing the maintenance cycles, and eventually prolonging the structural life with minimized cost. However, for most structures of practical significance, probability density evolution is usually governed by a high-dimensional nonlinear partial differential equation (PDE) whose solution is computationally expensive or impractical for optimization or real-time applications.

Considering the random event description of the principle of preservation of probability, Li and Chen proposed an effective and novel analysis method called Probability Density Evolution Method (PDEM) [1, 2]. PDEM can be used to analyze stochastic structures. In addition, by deriving the General Density Evolution Equation (GDEE) they proved that in stochastic structural systems, the equations governing the evolution of the probability density can be decoupled by introducing the physical solution of the system [3]. Over the past decade, PDEM has been used successfully in many numerical studies on the stochastic behavior of the structures [4-6] and has been validated in experimental studies [7, 8]. Although PDEM decouples the probability density evolution PDE in the probability space [9], numerical treatment of the governing equations is still stringent and impractical for real-time applications.

Fortunately, recent significant growth and advancements in machine learning have highlighted new methods to solve the probability density evolution via machine learning and deep networks. In this respect, Zhou and Peng [10] proposed the utility of active learning and Gaussian process regression to estimate and solve a high-dimensional probability density evolution equation. Haj and Soubra used the Kriging method and Monte-Carlo Simulation (MCS) to assess the failure probability of a structure [11]. Yang et al. used polynomial regression and Support Vector Machine to estimate structural life-cycle reliability [12]. Machine learning in general and deep neural networks, in particular, are known to be highly efficient at modeling nonlinear functions of high-dimensional inputs and a compelling tool for exploring hidden correlations.

More recently, Raissi et al. used prior physical knowledge and the conservation laws to define the loss function of a deep neural network (DNN) by proposing Physics-Informed Neural Networks (PINNs) to solve PDEs [13]. Taking advantage of the Automatic Differentiation (AD), PINNs can represent all differential operators in a given PDE. Unlike many other numerical methods, PINNs are grid-free [14] that can reduce the required computational effort significantly. Therefore, PINNs present a generalized procedure to solve and generate a surrogate for the solution of PDEs regardless of their structure.

As indicated in [15] the main challenges faced in surrogate modeling are Data Efficiency, Uncertainty Quantification, and Generalization. When real-life applications are considered, the problem is usually high-dimensional, where the computational efficiency of the probability density evolution method is still known to be formidable due to the curse of dimensionality [16]. Therefore, to find the density evolution, we still need to solve a couple of thousands of computationally expensive structural analyses. A possible remedy is to use surrogates such as Gaussian Process Regression (GPR), Polynomial Chaos Expansion, etc. However, the accuracy and efficiency of these surrogates are highly dependent on the dataset used to construct them. In other words, these techniques do not use the physical constraints and relations between the training datasets to construct the surrogates. Therefore, they either require a lot of sample points to provide an accurate surrogate or need to be tuned specifically for the problem that is being solved so that they can provide an adequate accuracy with a relatively small training dataset [17]. Generalizability is another challenge for surrogates. The generalizability of the DNNs in general and the PINNs, in particular, is an ongoing research subject. However, it is known empirically that deep networks exhibit good generalization behavior [18]. This generalization potential can provide a unique opportunity for solving probability density evolution in real-time applications.

In this research, the most recent studies and developments in Machine Learning (ML) are considered to propose a physics-informed deep learning algorithm to assess the probability density evolution of a probability conservative stochastic structure. In this respect, a Latin hypercube sampling strategy is used to generate the sample points in the spatiotemporal space at each iteration in the training process. This way, the mesh forming is avoided and PDEM is converted to an ML problem. Subsequently, the neural network is trained such that it satisfies both the Initial Condition (IC) and the differential operator. Next, the PINNs concept is utilized to encode the GDEE dynamics into the network as well providing a deep physics-informed network (DeepPDEM) that not only solves the density evolution problem but also can be used as a surrogate for the GDEE. Since DeepPDEM uses the physical insight in its loss function, it requires a smaller dataset to train [19]. It is worth noting that the proposed method does not rely on labeled training data sets and the network learns the GDEE through randomly selected points from the domain. Application of the proposed scheme ensures that the network is an eligible surrogate that does not over-fit the training data set. The efficacy of the proposed approach is tested through several numerical experiments. Since it is usually of interest to solve the GDEE over a range of physical setups and initial conditions, a few experiment setups with analytical solutions are used for benchmarking, verification, and to demonstrate the applicability of the proposed concept. DeepPDEM is used to approximate the probability density evolution of these experiment setups, with different initial and physical conditions whose results are compared with exact analytical solutions.

## 2 Generalized Probability Density Evolution

Given the probability density distribution at some instance of time for a dynamic system, one can use the system's equation to find the evolution of the probability density for a future instant of time. In this study, the probability-conservative or preserving structures are considered. Probability conservation states that the probability of a random factor involved in a stochastic process is preserved, in other words, no random factors arise or vanish during the process. Considering Noether's first theorem, every differentiable symmetry of the action of a physical system has a corresponding conservation law. A local probability conservation law can be expressed as a continuity equation, a PDE that gives a relation between the amount and the transport of the probability. The continuity equation gives the most general form of an exact conservation law. The conservation law is usually expressed using a quantity, that can move or flow, and its volume density. Assuming the quantity of interest to be the probability (P), its volume density is naturally the Probability Density Function (PDF) ($\wp$). Think of the probability evolution as a flow of probability that can be described by probability flux. The probability flux, which is denoted by $\mathbb{j}$, is a vector field. In some research areas such as quantum mechanics, the probability flux is also known as the probability current. The dimension of probability flux is the "amount of probability flowing per unit of time, through a unit area $[s^{-1}m^{-2}]$".

Considering a velocity vector field $\mathbf{u}$, if $\mathbf{u}(\mathbf{x})$ describes the velocity with which the probability at point $\mathbf{x}$ is moving, then by definition, the probability flux is equal to the PDF times the velocity vector field.

$$\mathbb{j} = \wp_{\mathbf{X}}(\mathbf{x}, t)\mathbf{u} \tag{1}$$

Thus, for the continuity equation:

$$\frac{\partial \wp_{\mathbf{X}}(\mathbf{x}, t)}{\partial t} + \nabla \cdot (\wp_{\mathbf{X}}(\mathbf{x}, t)\mathbf{u}) = \mathscr{s} \tag{2}$$

where $\mathscr{s}$ represents a source term, showing the generation of probability per unit volume per unit time, and "$\nabla \cdot$" is the divergence operator. Since the PDF is a scalar-valued function, one could write;

$$\frac{\partial \wp_{\mathbf{X}}(\mathbf{x}, t)}{\partial t} + \left(\nabla \wp_{\mathbf{X}}(\mathbf{x}, t)\right) \cdot \mathbf{u} + \wp_{\mathbf{X}}(\mathbf{x}, t)(\nabla \cdot \mathbf{u}) = \mathscr{s} \tag{3}$$

Equation (3) represents the probability density evolution equation. As stated earlier, we are interested in probability preserving systems in this research. We may neglect the source term and thus consider the following form:

$$\frac{\partial p_{\mathbf{X}}(\mathbf{x},t)}{\partial t} + \left(\nabla p_{\mathbf{X}}(\mathbf{x},t)\right) \cdot \mathbf{u} + p_{\mathbf{X}}(\mathbf{x},t)(\nabla \cdot \mathbf{u}) = 0 \qquad (4)$$

This equation can be further simplified considering the dynamics of a time-invariant structure. It is known that the evolution of a dynamic system preserves the probability if the randomness involved is kept unchanged [2]. This essentially means that by keeping the random events unchanged, the probability is preserved in a mathematical transformation. Let "**η**" and "**F**" denote random parameters with known joint PDFs and stochastic excitation, respectively. The Equation of Motion (EOM) for a stochastic second-order dynamic can be written as:

$$\mathbf{M}(\boldsymbol{\eta})\ddot{\mathbf{X}} + \mathbf{C}(\boldsymbol{\eta})\dot{\mathbf{X}} + \mathbf{g}(\boldsymbol{\eta}, \mathbf{X}) = \mathbf{F}(t) \qquad (5)$$

where **C** and **M** are the damping and the mass matrices, respectively. Further, assume "**F**" to be a square-integrable zero-mean random process with continuous covariance defined over a bounded and closed interval $\varpi$. Via the utility of the Kosambi-Karhunen-Loève (KKL) theorem [9] to decompose the stochastic forcing function, an optimal series expansion in the MSE sense can be proposed for a square-integrable zero-mean random process. Using the eigenvalues and eigenfunctions of the autocovariance operator describing the random field, the KKL theory employs uncorrelated random coefficients to expand the random process. Let $\zeta_k$ be the uncorrelated random variables with known joint PDF defined as:

$$\zeta_k = \int_{\varpi} \mathbf{F}_i(t) e_k(t) \mathrm{d}t \qquad (6)$$

where $e_k$ shows an orthonormal basis on $L^2(\varpi)$.

$$\mathbf{F}_i(t) = \sum_{k=1}^{\infty} \zeta_k e_k(t) = \mathbf{F}_i(\boldsymbol{\zeta}, t) \qquad (7)$$

Now, allow "**Θ**" to show a vector of random variables consisting of both $\eta$ and $\zeta$. Thus, Eq. (5) can be rewritten as:

$$\mathbf{M}(\boldsymbol{\Theta})\ddot{\mathbf{X}} + \mathbf{C}(\boldsymbol{\Theta})\dot{\mathbf{X}} + \mathbf{g}(\boldsymbol{\Theta}, \mathbf{X}) = \mathbf{F}(\boldsymbol{\Theta}, t) \qquad (8)$$

One could consider a function of the states to find any other physical quantity or response of interest from the structure. Let "$\mathbf{X}(\boldsymbol{\Theta}, t)$" and "$p_{\mathbf{X}\boldsymbol{\Theta}}(\mathbf{x}, \boldsymbol{\theta}, t)$" represent the time evolution of the response as a function of $\boldsymbol{\Theta}$ and the instantaneous joint PDF of "$\mathbf{X}$" and "$\boldsymbol{\Theta}$" at a time "$t$". With this definition, the Generalized Probability Density Evolution Equation (GDEE) can be used to find the probability density evolution of the structure [20]:

$$\begin{cases} \dfrac{\partial p_{\mathbf{X}\boldsymbol{\Theta}}(\mathbf{x}, \boldsymbol{\theta}, t)}{\partial t} + \sum_{l=1}^{m} \dot{X}_l(\boldsymbol{\theta}, t) \dfrac{\partial p_{\mathbf{X}\boldsymbol{\Theta}}(\mathbf{x}, \boldsymbol{\theta}, t)}{\partial x_l} = 0 & \mathbf{x} \in \Omega_{\mathbf{x}}, \boldsymbol{\theta} \in \Omega_{\boldsymbol{\theta}}, t \in [0, T] \\ \mathcal{I}(\mathbf{x}, \boldsymbol{\theta}) = p_{\mathbf{X}\boldsymbol{\Theta}}(\mathbf{x}, \boldsymbol{\theta}, t)|_{t=t_0} = \delta(\mathbf{x} - \mathbf{x}_0) p_{\boldsymbol{\Theta}}(\boldsymbol{\theta}) \end{cases} \quad (9)$$

$$p_{\mathbf{X}}(\mathbf{x}, t) = \int_{\Omega_{\boldsymbol{\theta}}} p_{\mathbf{X}\boldsymbol{\Theta}}(\mathbf{X}, \boldsymbol{\theta}, t) d\boldsymbol{\theta} \quad (10)$$

In equation (9), $\dot{X}_l(\boldsymbol{\theta}, t)$ are functions of the independent variables only. Therefore, GDEE is a first-order linear hyperbolic PDE with the Dirichlet boundary condition. The method of characteristics can be employed to solve this equation. This equation can be solved numerically once the coefficients $\dot{X}_l(\boldsymbol{\theta}, t)$ are known. Therefore, to solve the GDEE, one can first choose a sample set in the random parameter space $\Omega_{\boldsymbol{\theta}}$. Then, for each of these points, solve the GDEE separately. This leads to an ensemble of trajectories in the probability space that represents the probability density evolution. So, let's assume that the probability density function $p_{\mathbf{X}\boldsymbol{\Theta}}(\mathbf{x}, \boldsymbol{\theta}, t)$ can be well-approximated by a polynomial of degree $2n - 1$ or less on its domain of definition $\Omega_{\boldsymbol{\theta}}$. Given that the domain $\Omega_{\boldsymbol{\theta}}$ is bounded, the integral in Eq.(10) can be approximated by Gauss-Legendre quadrature.

## 3 Deep Learning Estimation of the GDEE

In this section, a DNN algorithm is proposed to approximate the time evolution of the probability of a stochastic structure using GDEE. Later, the proposed algorithm is used to solve several problems with different physical and boundary conditions.

DNNs are universal approximators [21], nonlinear composition functions in which the inputs are transformed with linear and nonlinear mappings recursively. DNNs provide a systematic and powerful approach to identifying correlations in data. This advantage is achieved with a high computational cost for the training process. However, in most cases, the training can be handled offline with the existing powerful parallel computing tools.

To estimate the solution by a DNN, let "$f$" denote an approximation of the instantaneous PDF found by the neural network $\mathcal{N}$. Further, let "$\mathcal{L}$" represent the approximation error function (loss function) representing the $L^2$-norm of the PDE and the IC residuals that measures how well the differential operator and the initial conditions are approximated. A key step to approximate the solution of GDEE with a deep network is to constrain it to minimize the GDEE in residual form. So, one could write,

$$\mathcal{N}^L(\mathbf{x}, t; \boldsymbol{\chi}) = f\left(\mathbf{x}, t, \frac{\partial \tilde{p}}{\partial t}, \frac{\partial \tilde{p}}{\partial z}; \boldsymbol{\chi}\right) \quad \mathbf{x} \in \Omega_{\mathbf{x}}, t \in [0, T] \qquad (11)$$

$$\mathcal{L}(f) = \alpha_1 \left\|\frac{\partial f(\mathbf{x}, \boldsymbol{\theta}, t; \boldsymbol{\chi})}{\partial t} + \sum_{l=1}^{m} \dot{X}_l \frac{\partial f(\mathbf{x}, \boldsymbol{\theta}, t; \boldsymbol{\chi})}{\partial x_l}\right\|_2^2 \\ + \alpha_2 \|f(\mathbf{x}, \boldsymbol{\theta}, 0; \boldsymbol{\chi}) - \delta(\mathbf{x} - \mathbf{x}_0) p_{\boldsymbol{\theta}}(\boldsymbol{\theta})\|_2^2 \quad \boldsymbol{\theta} \in \Omega_{\boldsymbol{\theta}} \qquad (12)$$

$\alpha_1$ and $\alpha_2$ are strictly positive normalizing coefficients. "$L$" represents the number of network layers and finally $\boldsymbol{\chi} = [\mathbf{W}^\ell, \mathbf{b}^\ell]_{1 \leq \ell \leq L}$ are the parameters (weights and biases) of the neural network. Here we assume no knowledge of the actual instantaneous probability density. By minimizing the loss function "$\mathcal{L}$" to be as close to zero as possible, the approximation "$f$" approaches the instantaneous PDF and closely satisfies the differential equation and the initial condition. Ideally, if $\mathcal{L}(f) = 0$, then the approximation is the exact solution of the GDEE.

In previous research on PINNs, Raissi et al. [13] revisited the idea of using a PDE in residual form as a loss function for DNN to find and approximate the solution. As a result, by choosing the trainable parameters such that the loss function is minimized, the PDE is encoded into the network.

The fact that PINNs do not need any discretization scheme to solve a PDE distinguishes them from traditional complicated numerical solvers. PINNs rely on Automatic Differentiation (AD) to satisfy the differential operators in a PDE, a capability that was employed for a very long time in neural networks and backpropagation. Leveraging from the fact that the DNNs can be regarded as universal function approximators [22], a DNN is trained to capture the nonlinear mapping from the input space to the output space that satisfies the boundary and differential equation of interest. Therefore, the network relies on the physics described by the PDE and provides a nonlinear approximation of both the solution as well as its derivatives.

As stated earlier, we would like to propose a data-driven solution for the GDEE PDE and approximate its solution with a PINN instead of traditional numerical methods. Since the PINNs obtain the solution by generally solving a non-convex optimization problem, a unique solution cannot be guaranteed as opposed to conventional numerical methods [14]. Therefore, let's assume that:

1. There exists a unique solution for the GDEE of interest,
2. The solution is bounded,
3. The solution is uniformly Lipchitz and changes continuously with the initial condition [23],
4. The architecture and the number of neurons/layers used are sufficient to approximate the desired solution,
5. A finite set of collocation points can be used to estimate the loss of the network with sufficient accuracy (Eq. (12)).
6. The number of collocation points chosen is sufficiently large that the estimated network loss is a good representative of the approximation error [13].

Let $n_{\partial\Omega}$ and $n_\Omega$ show the number of arbitrary collocation points chosen on the boundary (the initial condition) and within the domain of interest, respectively. Therefore, we can approximate the loss function of the neural network on these collocation points, given that the derivatives of the output with respect to the spatiotemporal inputs can be computed.

$$\mathcal{L}_\Omega(\mathcal{N}) = \frac{1}{n_\Omega} \sum_{i=1}^{n_\Omega} \left\| \frac{\partial \mathcal{N}(\mathbf{x}^i, t^i)}{\partial t^i} + \sum_{l=1}^{m} \dot{X}_l \frac{\partial \mathcal{N}}{\partial x_l} \right\|_2^2 \tag{13}$$

$$\mathcal{L}_{\partial\Omega}(\mathcal{N}) = \frac{1}{n_{\partial\Omega}} \sum_{j=1}^{n_{\partial\Omega}} \left\| \mathcal{N}(\mathbf{x}^j, 0) - \delta(\mathbf{x}^j - \mathbf{x}_0) \wp_{\boldsymbol{\theta}}(\boldsymbol{\theta}) \right\|_2^2 \quad \boldsymbol{\theta} \in \Omega_{\boldsymbol{\theta}} \tag{14}$$

$$\mathcal{L}(\mathcal{N}) = \frac{\mathcal{L}_\Omega(\mathcal{N})}{\mathcal{L}_\Omega(\mathcal{N}^0)} + \frac{\mathcal{L}_{\partial\Omega}(\mathcal{N})}{\mathcal{L}_{\partial\Omega}(\mathcal{N}^0)} \tag{15}$$

where $\mathcal{L}(\mathcal{N}^0)$ denotes the initial loss of the network. To compute the derivatives of the estimated solution provided by the net with respect to the spatiotemporal inputs, four possible methods can be considered [24] that include: (1) numerical approximations, (2) symbolic differentiation, (3) analytic differentiation, and (4) automatic differentiation. Using backpropagation, AD was used in neural networks almost since its inception [25].

Since all the computations within a neural network model are finite compositions of elementary operations with known derivatives, AD uses chain rule to differentiate the overall composition. Unlike numerical schemes such as finite-difference in which computing partial derivatives requires at least two function evaluations along each dimension, using the chain rule lets AD determine the derivatives independent of the dimensions. This allows us to avoid numerical and truncation errors while computing the derivatives more accurately and efficiently. Furthermore, AD can also be employed recursively to compute the $n^{th}$-order derivative of the network as well, and again, since the derivation is already taken care of with AD, noisy data and other similar numerical instabilities do not raise an issue [26].

AD is also implemented in well-known deep learning open-source software libraries, such as TensorFlow [27] and PyTorch [28]. In this study, we use TensorFlow to implement the network model and therefore benefit from the AD implementation in this software package to find the partial derivatives of Eq. (13).

So far, we have not imposed any restrictions on the collocation points that are used to train the neural network except that their number should be sufficiently large that the estimated network loss is a good representative of the approximation error. This is almost like applying Euler discretization on Eq. (12). Therefore, one can use uniformly sampled random points to determine the collocation points whether inside the spatiotemporal domain or on the boundaries [13, 29]. As a result, DeepPDEM is a mesh-free solver. Once the collocations points are determined, we can use an optimizer to find the parameters $\chi$ such that the loss function is minimized. In this study the networks are trained for a predetermined number of epochs, whose convergence are recognized if and only if (IFF) for the last ten percent of the total epochs the relative decrease in the loss is less than 0.01 percent. Figure 1 shows the solution algorithm schematically and the flowchart is depicted in Figure 2.

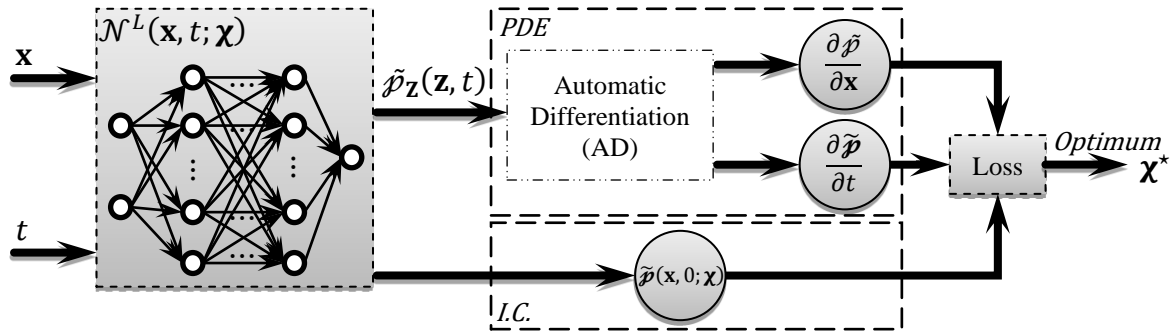

Figure 1- Solution procedure of the GDEE using a DNN

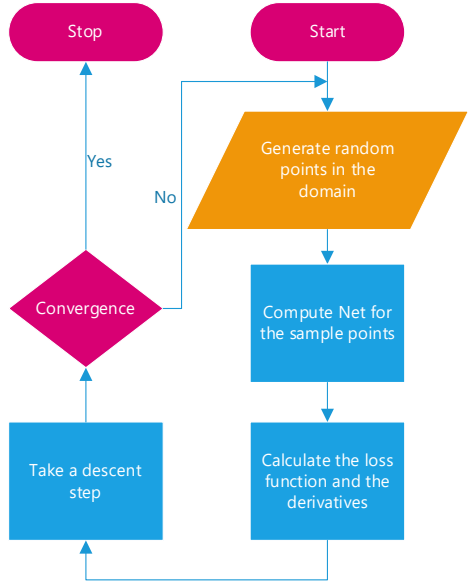

Figure 2- Solution flowchart

## 3.1 Choosing Collocation Points

In their study on the tradeoffs of large-scale learning, Bottou and Bousquet [30] decomposed the error of a neural network into (a) approximation error $\mathcal{E}_{app}$, (b) estimation error $\mathcal{E}_{est}$, and (c) optimization error $\mathcal{E}_{opt}$. They recognized approximation error as a measure that how closely the family of functions chosen to construct the network can approximate the solution. They also stated that the estimation error, on the other hand, estimates the effect of minimizing the empirical risk instead of the expected risk (as a result of enforcing the Euler discretization in the loss function), while the optimization error arises from the fact that the empirical risk optimization itself is usually computationally intensive and sometimes the iteration in DNNs training is stopped before convergence. In other words, to improve the approximation error one can suggest changing hyperparameters such as the number of neurons/layers or activation functions. On the other hand, collocation points, their count, and location play an important role to improve the estimation error.

The estimation error can also be a measure of overfitting. A possible solution to decrease the chance of overfitting is to change the collocation points in each iteration. Since DeepPDEM is a mesh-free solver, resampling can be easily carried out on the boundary and interior as the algorithm progresses to avoid overfitting. However, random choice of the collocation points with no strategy may not always offer good performance. In this study, we particularly focus on cases where the solution exhibits steep gradients as this is the case that arises in the examples solved in the current study.

As the training progresses, we can assume that the gradient estimated by the network approaches the solution gradient. We suggest using the estimated gradient to enhance the sampling. The collocation points in the interior of the spatiotemporal domain are divided into two groups, the first group of sample points is generated using an LHS strategy [31] with uniform distribution. Similar to the concept of importance sampling [32], the second group is sampled with an importance distribution proportional to the estimated gradient.

The idea of importance sampling is utilized to enhance the performance of the algorithm. To this end, the estimated gradient of the solution is used. Let $\nabla p_{\mathbf{X}}(t, \mathbf{x})$ denote the estimated gradient vector of the solution. An importance density $q(t, \mathbf{x})$ can now be defined as the inner product of the gradient vector:

$$q(t, \mathbf{x}) = \frac{1}{c} \langle \nabla p_{\mathbf{X}}(t, \mathbf{x}), \nabla p_{\mathbf{X}}(t, \mathbf{x}) \rangle^{\frac{1}{2}} \tag{16}$$

where $c$ is the normalization constant. In each iteration, this distribution is then used by the importance sampling algorithm to select the second group of the sample points. Therefore, whenever the solution has a steep gradient (higher density value), it is more probable that the algorithm adds more collocation points. Given more data with this mechanism, the algorithm can regulate the loss function and estimate the output with better accuracy.

## 3.2 LSTM-Based PDEM

Since Multi-Layer Perceptron (MLP) is sufficient for dealing with most PDEs [13] it is our primary choice to consider in this study. However, MLP is the simplest DNN architecture. Using other architectures might improve the results, or increase the convergence rate. Considering sequences and data with time dependency, Recurrent NNs (RNN) might also help capture the dependencies in the solution easier and counteract the estimation error. Long Short-Term Memory (LSTM) [33], a well-known recurrent network, uses internal states to capture long and short-term dependencies in the input data. In this study, we considered both MLP and LSTM and compared their performance.

Conventional recurrent network models, such as LSTM, are time-series analyzers utilizing past observations to generate a future prediction. The recurrent structures in an LSTM model can capture the temporal dynamics of the training data. Therefore, vanilla LSTM models work with regular constant sampling period time-series data. In this study, the vanilla LSTM model with a constant mesh over the spatiotemporal space is used. In this sense, for each iteration, a vector of the initial condition is fed to the network that is set to return a sequence as the output. The output sequence is then reshaped to form an output matrix that represents the solution over a fixed spatiotemporal mesh. Utilizing the multivariate cubic spline interpolation scheme over the rectangular spatiotemporal mesh, this output represents the response surface of the network which is sampled to evaluate the network loss.

## 3.3 Additional Notes and Training Procedure

Based on our experience, tuning all the hyperparameters of the network (such as initialization, learning rate, network dimension, etc.) is needed to achieve a solution with the best accuracy. However, these fine-tunings are case-dependent and usually cannot be generalized. Therefore, in this paper, we study the effects of network architecture, activation function, and sampling fraction hyperparameters on the accuracy of the solution.

As stated earlier, we consider two network architectures i.e., MLP and LSTM, to see whether RNNs can offer a better accuracy or not. Another hyperparameter that potentially has a significant effect on training dynamics and network performance is the activation function [34]. Traditional activation functions, such as the hyperbolic tangent, encounter vanishing gradient issues. Swish [34] is a non-monotonic, smooth function that helps alleviate the vanishing gradient problem and thus helps to improve the convergence rate of the network. Thus, the Swish activation function is used in an MLP network (MLP-Swish) to study the anticipated effects. The sampling fraction determines the ratio of spatiotemporal sample points that the algorithm chooses according to the importance sampling scheme, explained earlier, to better capture the solutions with a steep gradient. We also study the performance of training for both L-BFGS and Adam optimizers.

Moreover, Glorot normal initializer [35] was utilized in our networks to initialize the biases and the weights. As mentioned earlier, the training of DeepPDEM is a non-convex and nonlinear optimization problem. Therefore, different initial values for the trainable variables may result in different results for the trainable parameters [36] and different convergence rates [23]. As discussed in [37] Batch Normalization (BN) is a possible solution to decrease the sensitivity of the network with respect to the initialization. Batch Normalization can also improve the convergence of networks [38]. In our case, we chose the batch size of 200. No rescaling and centering schemes are used.

Even though using a gradient-based optimizer to train a deep network is inevitable given the number of trainable variables, it increases the chance of converging to local (and sometimes bad) optima. One possible strategy to avoid premature convergence in stochastic gradient descent algorithms such as Adam is to use an adaptive scheme for the learning rate [39, 40]. By keeping the learning rate small for the first few epochs of training and then increasing it gradually we can achieve a good balance between exploration and exploitation in the solution space. This highlights the importance of the training rate and its significant effect on the training and convergence rate [41]. In this respect, the rectified adaptive learning rate [40] was used to update the learning rate in the training process. Technically, as explained earlier we would like the learning rate to be as large as possible in the beginning and gradually decreases in the process. Therefore, we chose the highest value to start from in the adaptive learning rate algorithm such that none of our networks diverge, i.e., 0.0015. To compare the networks, we used a fixed number of iterations/epochs and report the results obtained. We also preferred to use Nondimensionalized Elapsed Time (N.E.T.) to compare the networks. Therefore, to find the N.E.T. we used the MLP-LBFGS elapsed time as the reference and divided the elapsed times with this value for each case. Other than what was explained earlier, TensorFlow was used with the recommended parameters to implement the networks.

## 4 Results and Analysis

### 4.1 Performance study

In this section three case studies are investigated that include: (1) Free vibration of an uncertain SDOF system, (2) Free vibration of an Euler-Bernoulli beam, and (3) Forced vibration of an Euler-Bernoulli beam. The case studies are specially chosen such that their analytical solution can be derived for verification purposes against the proposed DeepPDEM and to perform comparative studies better.

## 4.2 Implementation Details

Optimizing a DNN with respect to its loss function (training) is a non-convex and nonlinear problem. Heuristics approaches are often computationally intractable and impractical since there are a large number of trainable parameters. Gradient-based optimizers such as L-BFGS [42] and Adam [43, 44] are the de-facto methods of choice in machine learning, even though they might get stuck in a local minimum. Adding random parameters in the optimization process and using $1^{st}$-order derivatives of the loss function, Adam has better odds of finding the global optimum compared to L-BFGS which uses a quasi-Newtonian approach and thus $2^{nd}$-order derivatives of the loss function. Although, this comes at the cost of lower convergence rate. Despite all these to choose between these two optimizers, characteristics of the solution space for each problem shall be considered. Therefore, in this paper we used both optimizers and compared their performance.

As indicated before, the sampling fraction determines the ratio of spatiotemporal sample points that the algorithm chooses according to the importance sampling scheme, to the number of sampling points generated by the space-filling LHS. Therefore, to study the effects of this parameter, three different sampling ratios are chosen ({0,0.2,0.5}). Note that by definition a zero sampling ratio is equivalent to the ablation of the proposed importance sampling scheme.

Designing an optimal architecture for a DNN and optimizing its hyperparameters remains an open problem [45-47]. Pinkus showed that a Multi-Layer Perceptron (MLP) network with sufficient neurons can approximate a function and its partial derivatives [48]. Although, determining a sufficient number of neurons may not always be straightforward and may change from one problem to another. In this study, we used networks with a depth of 4 layers and a width of 20 neurons. Increasing the depth of the network increases the computational complexity of the network and thus decreases the convergence rate significantly. Even though some of the results presented in this paper can also be obtained via networks with fewer neurons and layers, we made sure that the chosen width and depth offer an adequate performance on all problems investigated in this paper. We found that the network depth has a significant role in overfitting and/or premature convergence.

The sensitivity of the solution with respect to network initialization is also studied. As stated earlier, the Glorot normal initializer [35] was used to initialize weights and biases. Therefore, the sensitivity of the solution to the weights initialization is first investigated. To this aim, the MLP-LBFGS network is used in 50 Monte-Carlo experiments. Utilizing different random seeds in each experiment, the network was initialized with the Glorot normal initializer [35] and the MLP-LBFGS network was trained for 50000 epochs. All other parameters were kept the same.

The learning rate shows a trade-off between exploration and exploitation in the optimizer. In this respect, the learning rate can be assumed as one of the most important hyperparameters to tune the training process. Selecting an adequate learning rate is usually performed via trial and error which could be a time-consuming process. As a remedy, a grid search strategy is utilized in the present study to check the sensitivity of the results w.r.t. the learning rate assumed for the rectified adaptive scheme. A grid of 10 logarithmically spaced points was selected for the learning rates. Subsequently, the MPL-Adam network was used and checked the loss after 50000 training epochs where three random seeds (1, 123123, 321321) were chosen to test the results.

**Case Study I: Response of an uncertain single-degree-of-freedom system**

Consider an uncertain single degree of freedom system as given in Eq. (17) [9]. In this equation, $\omega$ is assumed to be a random variable. Its probability density is uniformly distributed over the interval $[\omega_1, \omega_2]$.

$$\ddot{x} + \omega^2 x = 0 \; ; \; \dot{x}(0) = 0 \; ; \; x(0) = x_0 \tag{17}$$

Using the PDEM, the differential equation for the probability density evolution can be obtained as:

$$\frac{\partial p_{X,\Omega}(x,\omega,t)}{\partial t} - x_0 \omega \sin(\omega t) \cdot \frac{\partial p_{X,\Omega}(x,\omega,t)}{\partial x} = 0$$
$$p_{X,\Omega}(x,\omega,t_0) = \delta(x - x_0) p_\Omega(\omega) \tag{18}$$

Assuming $x_0 = 0.1$, $\dot{x}_0 = 0$, $\omega_1 = \frac{\pi}{4}$, $\omega_2 = \frac{3\pi}{4}$ and $t \in [0.9, 1.1]$, Eq. (18) can be solved whose details are provided in Appendix A.

$$p_X(x,t) = \frac{\left(\mathbb{H}\left(\frac{1}{t}\cos^{-1} 10x - \frac{\pi}{4}\right) - \mathbb{H}\left(\frac{1}{t}\cos^{-1} 10x - \frac{3\pi}{4}\right)\right)}{\frac{\pi}{2}|0.1 t \sin(\cos^{-1} 10x)|} \tag{19}$$

Next, we use the DeepPDEM framework to solve Eq. (18). In all test cases, we use the same hyperparameters as explained earlier. Figure 3 depicts the spatiotemporal sampled points of the last training epoch for the MLP-LBFGS network. As suggested in section 3.1, using the estimated gradient helps us calibrate the loss function and capture the regions with steep gradients better.

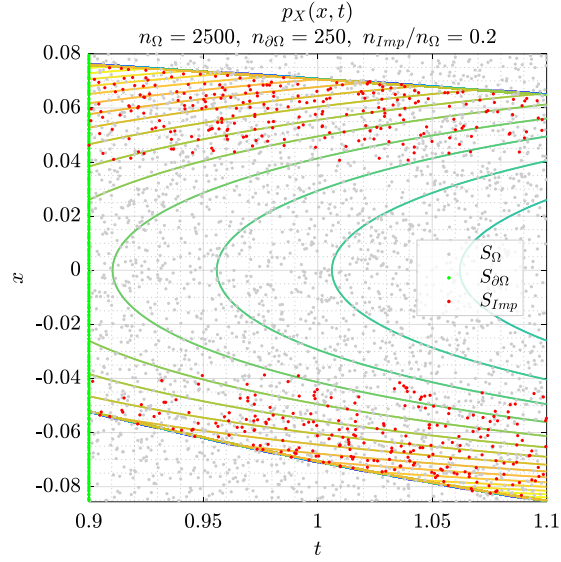

Figure 3. SDOF system: Three sets of randomly generated spatiotemporal collocation points are used to train the network, 1: LHS generated points, 2: IC points, and 3: importance-sampled points based on the predicted gradient

Table 1 summarizes the results of the first test case. As is indicated in this table, increasing the sampling fraction enhances the performance of the architectures using the Adam optimizer. Usage of the Swish activation function provides a better performance suggesting an easier solution space for the optimizer to explore.

Table 1 Case-Study (I) NN Depth = 4, NN Width = 20, Collocation Points = 2500, Epochs = 50000

| Network | Sampling Fraction = 0.0 | | Sampling Fraction = 0.2 | | Sampling Fraction = 0.5 | |
|---|---|---|---|---|---|---|
| | Loss | N.E.T. | Loss | N.E.T. | Loss | N.E.T. |
| MLP-LBFGS | 0.982246 | 0.778301 | 0.665870 | 1* | 0.901322 | 1.186415 |
| LSTM-LBFGS | 0.347042 | 1.158662 | 0.267899 | 1.471415 | 0.400199 | 1.801917 |
| MLP-Adam | 0.163257 | 0.725468 | 0.234631 | 0.920916 | 0.139924 | 1.260257 |
| LSTM-Adam | 0.141951 | 1.089507 | 0.214326 | 1.304675 | 0.094818 | 1.676997 |
| MLP-Swish-LBFGS | 0.035553 | 0.852861 | 0.017471 | 1.120684 | 0.005105 | 1.313472 |

(*) t=53596 s

Figure 4 shows the instantaneous PDF found by each network at $t = 1$. Figure 5 illustrates the estimation error with respect to the exact solution. It is realized that the MLP-Swish-LBFGS network shows the best performance and has been able to capture the density value near the boundaries more accurately.

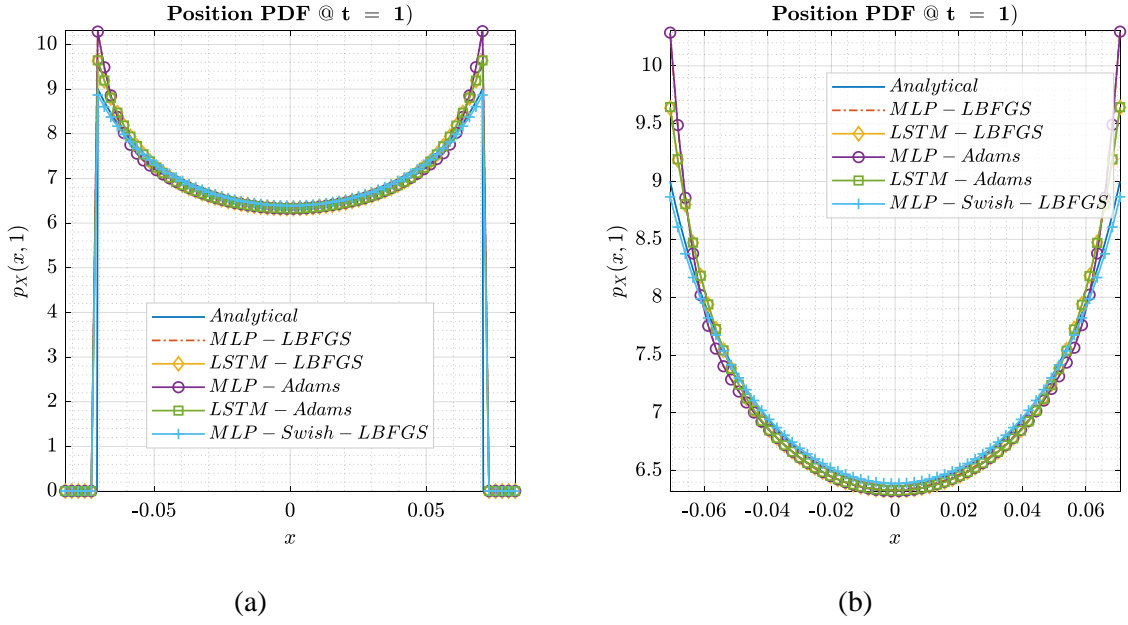

Figure 4. SDOF system: Comparison of DeepPDEM solutions with the exact solution at t = 1.
(a) Overall behavior of the PDF, (b) PDF behavior with scaled up vertical axis limits

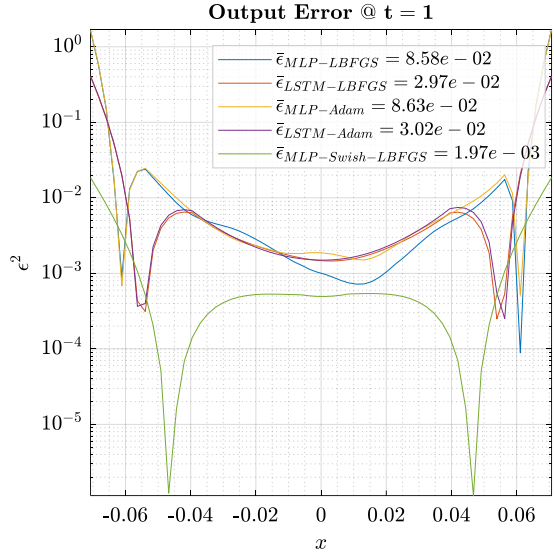

Figure 5. SDOF system: Comparison of DeepPDEM estimation error with respect to the exact solution at t = 1

- *Sensitivity Analysis*

Figure 6 represents the sensitivity of the loss function w.r.t. the weights initialization. The results indicate a positively skewed distribution ($G_1 = 5.0262$). The sample mean is $\mu = 0.6324$ and the sample variance is $S^2 = 0.0177$. Thus, it can be concluded that on average the output does not show strong sensitivity to the initialized weights.

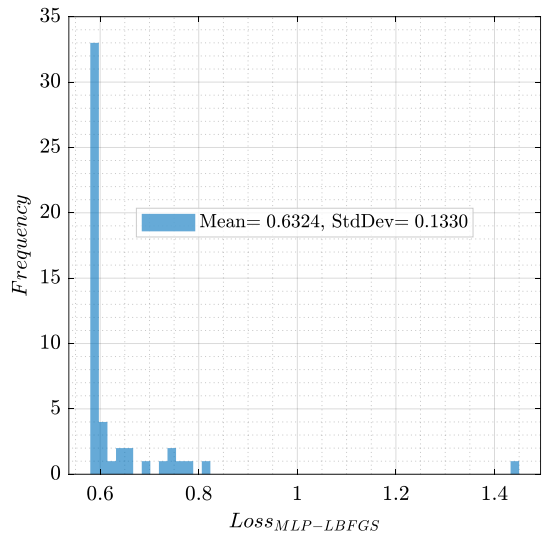

Figure 6. SDOF system: variation of DeepPDEM loss with respect to Glorot normal initializer using different random seeds in 50 Monte-Carlo experiments

Figure 7 presents the results obtained from the learning rate study. Smaller learning rates indicate more exploitation and greater ones indicate more exploration. Thus, it is expected that a network with a greater learning rate converges faster but possibly to a local optimum and a network with a smaller learning rate requires more training epochs to reach the solution. As the outcome shows, adequate results can be expected for the learning rates in the interval [0.001,0.015].

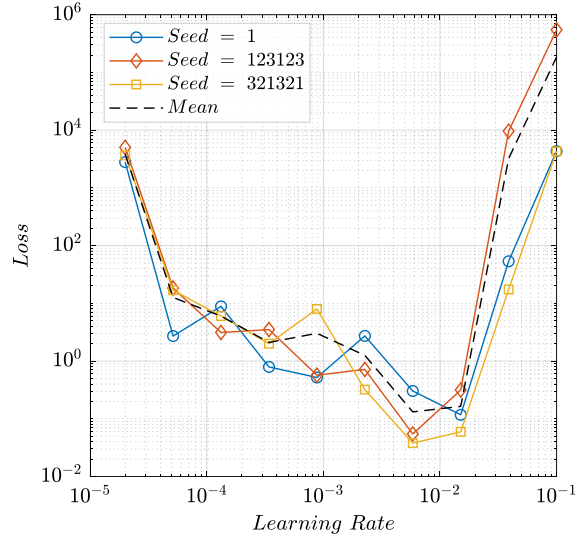

Figure 7. SDOF system: variation of DeepPDEM loss with respect to different rectified adaptive learning rates considering three different seeds

- *Generalization Error and Data Efficiency*

To check the data efficiency, the generalization error was used as the figure of merit and DeepPDEM was compared with two well-established surrogates, i.e., vanilla feed-forward shallow network and Gaussian Process Regression (GPR). To keep the analysis simple, only the MLP-LBFGS network was considered while the sampling fraction was set to zero. All surrogates are trained by identical batches of 400, 800, 1200, 1600, and 2000 randomly selected collocation points. The exact solution for the collocation points is used to train both the feed-forward shallow network and the GRP. Both the DeepPDEM and the shallow network are trained for 5000 iterations regardless of the loss function gradient value. For the shallow network, a network with 20 neurons was utilized and trained via the Levenberg-Marquardt algorithm. A constant basis function is used for the GPR surrogate. To check the random sampling effect on the performance, five consecutive simulations with different pseudorandom seeds are selected. To find an estimation of the generalization error a constant quadrilateral mesh over the solution space is considered (linear spacing with 1e-4 step) and an error matrix is defined between the exact solution and the predictions. For each surrogate, the maximum singular value of the error matrix was utilized to compare the results.

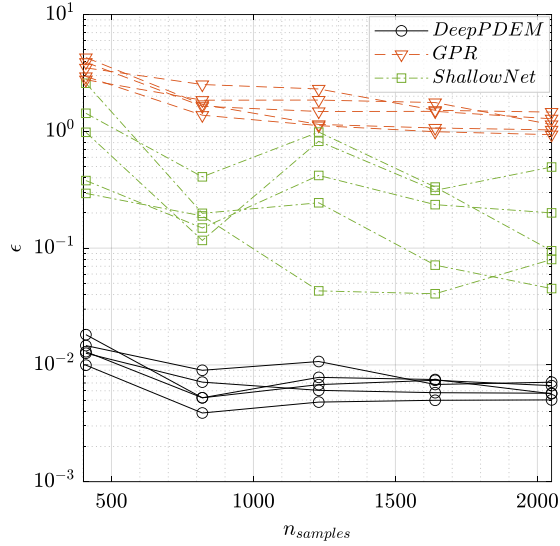

Figure 8 DeepPDEM, GPR, and Shallow NN surrogates' generalization error comparison with randomly selected training collocation points

Figure 8 presents a comparison of the performances among surrogates. As the results indicate, DeepPDEM (MLP-LBFGS) outperforms both of the traditional surrogates in terms of generalization accuracy. It is also evident that, as the DeepPDEM is using the underlying dynamics to construct the surrogate, it can achieve a better generalization error by using only 400 collocation points.

**Case Study II: Free Vibration of Euler-Bernoulli Beam**

Considering the flexural rigidity uncertainty to be uniformly distributed, the equations of motion of free vibration can be written as:

$$\begin{cases} \dfrac{\partial^2 w}{\partial t^2} + \alpha^2 \dfrac{\partial^4 w}{\partial x^4} = 0 \\ \\ w(0,t) = w_x(0,t) = 0 \quad ; \quad \alpha \sim U(\alpha_1, \alpha_2) \\ w_{xx}(l,t) = w_{xxx}(l,t) = 0 \\ \quad w(x,0) = \xi(x) \\ \quad w_t(x,0) = v(x) \end{cases} \quad (20)$$

Assuming $l = 1$, $0.9 \leq t \leq 1.1$, and $\alpha \sim U\left(\frac{\pi}{4}, \frac{3\pi}{4}\right)$, we can solve the GDEE for this system as detailed in Appendix B.

$$p_W(w, t) = \sum_\kappa \frac{\mathbb{H}\left(\alpha_\kappa - \frac{\pi}{4}\right) - \mathbb{H}\left(\alpha_\kappa - \frac{3\pi}{4}\right)}{\frac{\pi}{2}\left|0.8475t \cdot \cos\left(\alpha_k \mu_1^2 t + \frac{\pi}{4}\right)\right|} \quad (21)$$

Once again, we use the DeepPDEM framework to solve Eq. (20). Similar to the previous test case, the number of domain collocation points is chosen to be equal to 2500. The solution, in this case, is indicative of a discontinuous behavior that is captured by DeepPDEM. Figure 9 depicts the spatiotemporal sampled points of the last training epoch for the MLP-LBFGS network.

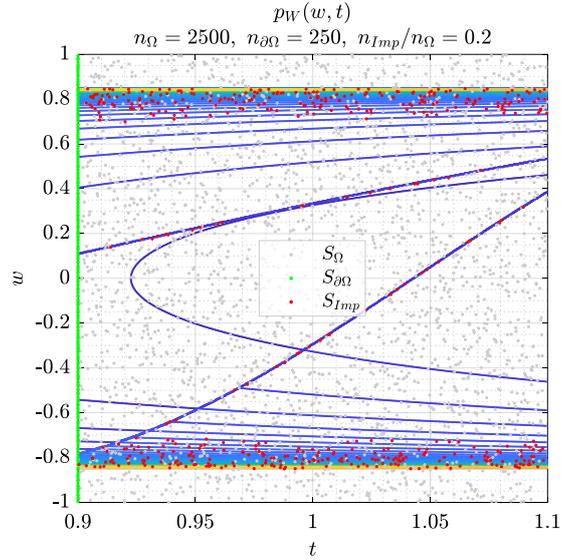

Figure 9. EB-beam free vibration spatiotemporal collocations: LHS-generated, IC, and importance-sampled points

Table 2 summarizes the results found for the second test case. In this scenario, the LSTM-LBFGS network shows the best accuracy, though the convergence time has increased significantly.

Table 2 Case-Study (II) NN Depth = 4, NN Width = 20, Collocation Points = 2500, Epochs = 50000

| Network | Sampling Fraction = 0.0 | | Sampling Fraction = 0.2 | | Sampling Fraction = 0.5 | |
|---|---|---|---|---|---|---|
| | Loss | N.E.T. | Loss | N.E.T. | Loss | N.E.T. |
| MLP-LBFGS | 0.022889 | 0.763048 | 0.020137 | $1^*$ | 0.001662 | 1.426825 |
| LSTM-LBFGS | 0.045192 | 1.253466 | 0.040313 | 1.543077 | 0.000490 | 2.167185 |
| MLP-Adam | 0.042952 | 0.554223 | 0.026095 | 0.751599 | 0.014583 | 0.938065 |
| LSTM-Adam | 0.003358 | 1.09985 | 0.005254 | 1.479458 | 0.000572 | 2.023670 |
| MLP-Swish-LBFGS | 0.034868 | 0.893117 | 0.005680 | 1.091968 | 0.027119 | 1.545853 |

$^{(*)}$ t=57686 s

We may again use the estimated instantaneous PDF at $t = 1$ to compare the results. Figure 10 and Figure 11 show the comparison between the five results against the exact solution. In this case, the MLP-Swish-LBFGS shows the poorest performance in capturing the discontinuity. The performance of this network degrades further by increasing the sampling fraction which can indicate a multi-modal solution space that causes the gradient-based optimizer to fail.

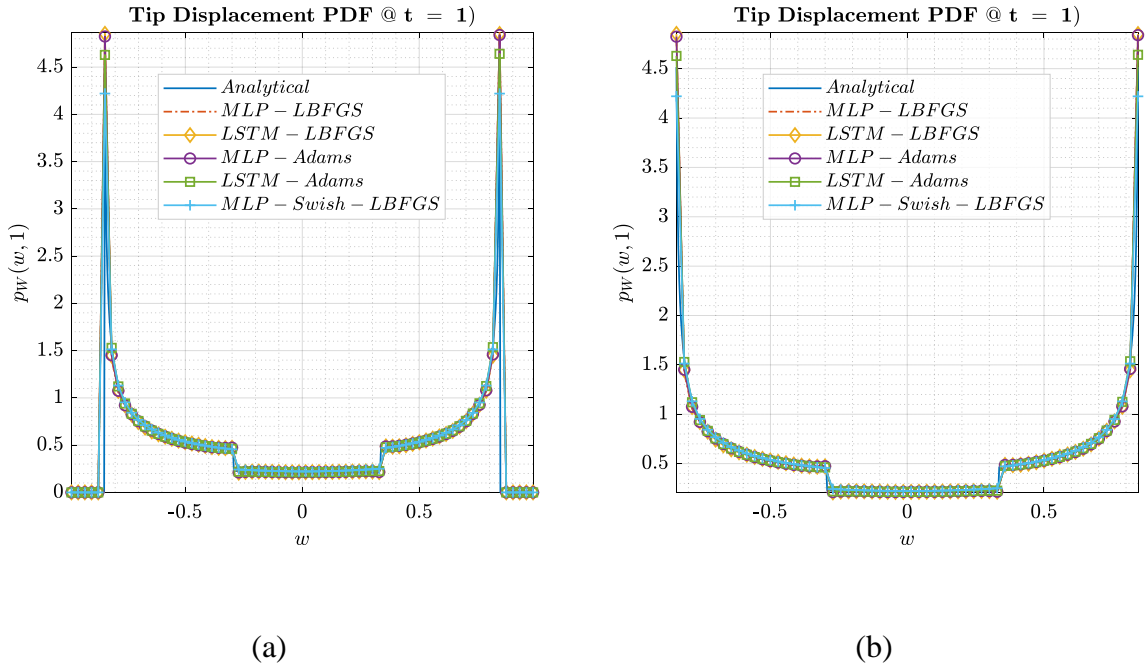

(a)    (b)

Figure 10. EB-beam free vibration: Comparison of DeepPDEM solutions with the exact solution at t=1. (a) Overall behavior of the PDF, (b) PDF behavior with scaled up vertical axis limits

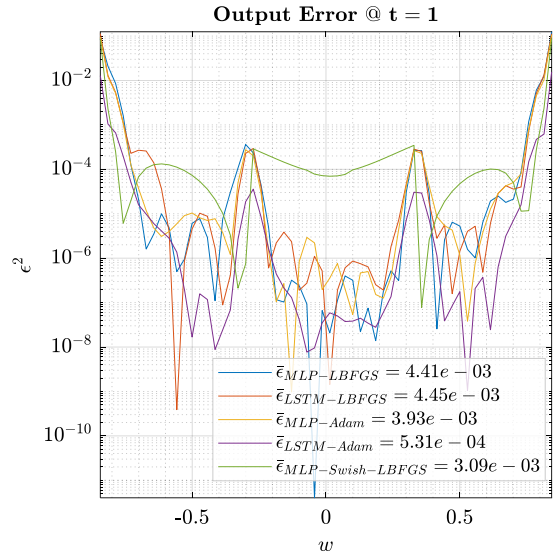

Figure 11. EB-beam free vibration: Comparison of DeepPDEM estimation error with respect to the exact solution at t = 1

- *Sensitivity Analysis*

The sensitivity of the loss function w.r.t. the weights initialization is given in Figure 12. In this test case, the skewness is $G_1 = 5.1946$, the sample mean is $\mu = 0.0730$ and the sample variance is $S^2 = 0.0396$. Therefore, on average, there is no strong sensitivity to the initialized weights.

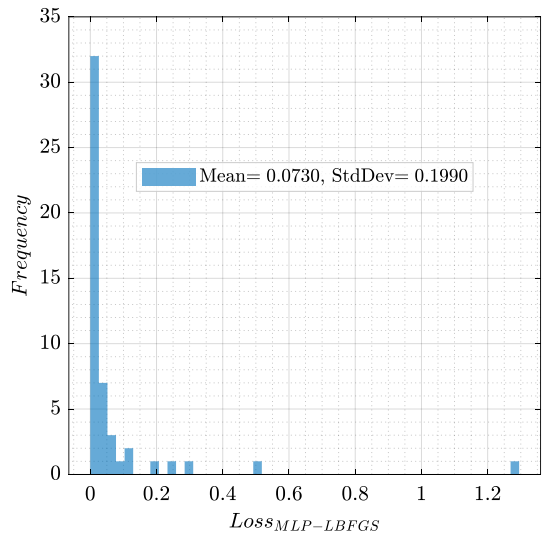

Figure 12. EB-beam free vibration: variation of DeepPDEM loss with respect to Glorot normal initializer utilizing different random seeds in 50 Monte-Carlo experiments

The algorithm diverged for the learning rate = 0.1 for the seeds 1 and 123123. The effect of the learning rate is presented in Figure 13 that indicates adequate results can be achieved for the learning rates within the interval $[1e-3, 1e-2]$.

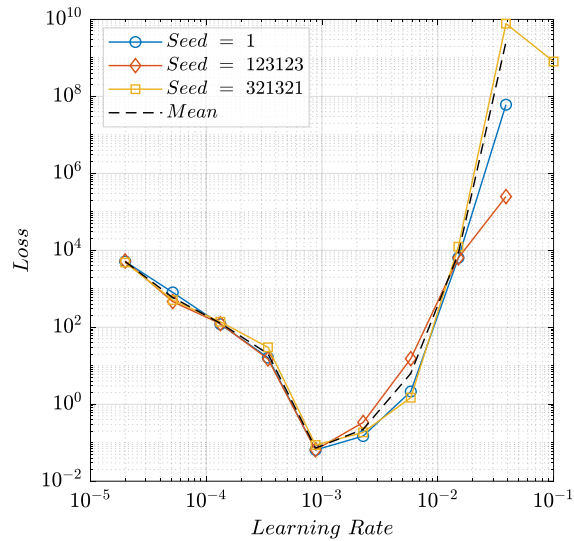

Figure 13. EB-beam free vibration: variation of DeepPDEM loss with respect to different rectified adaptive learning rates considering three different seeds

**Case Study III: Forced Vibration of Euler-Bernoulli Beam**

Considering the forcing function, the equation of motion is:

$$\begin{cases} \dfrac{\partial^2 w}{\partial t^2} + \alpha^2 \dfrac{\partial^4 w}{\partial x^4} = q_0 \sin(\Omega t + \psi)\, \phi_m(x) \\ \\ \quad w(0,t) = w_x(0,t) = 0 \qquad ;\quad \psi \sim U(\psi_1, \psi_2) \\ \quad w_{xx}(l,t) = w_{xxx}(l,t) = 0 \\ \quad w(x,0) = \xi(x) \\ \quad w_t(x,0) = v(x) \end{cases} \qquad (22)$$

The evolution of the probability for this system is governed by:

$$\begin{cases} \dfrac{\partial p_{W,\Psi}(w,\psi,t)}{\partial t} + \dot{W} \cdot \dfrac{\partial p_{W,\Psi}(w,\psi,t)}{\partial w} = 0 \\ I.C.:\quad p_{W,\Psi}(w,\psi,t_0 = 0) = \dfrac{\mathbb{H}(\psi - \psi_1) - \mathbb{H}(\psi - \psi_2)}{\psi_2 - \psi_1}\delta(w - w_0) \end{cases} \qquad (23)$$

Assume $m = 1$, $\alpha = 1$, $\Omega = 2$, $q_0 = 1$, and $l = 1$. The analytical solution for the probability density evolution of this system is derived in Appendix C and presented below in Eq. (24).

$$p_W(w,t) = \sum_\kappa \frac{\left(\mathbb{H}(\psi_\kappa - \psi_1) - \mathbb{H}(\psi_\kappa - \psi_2)\right)/(\psi_2 - \psi_1)}{|f'(\psi_\kappa)|} \qquad (24)$$

Using the DeepPDEM framework to solve Eq. (23) yields the spatiotemporal sampled points of the last training epoch for the MLP-LBFGS network as shown in Figure 14.

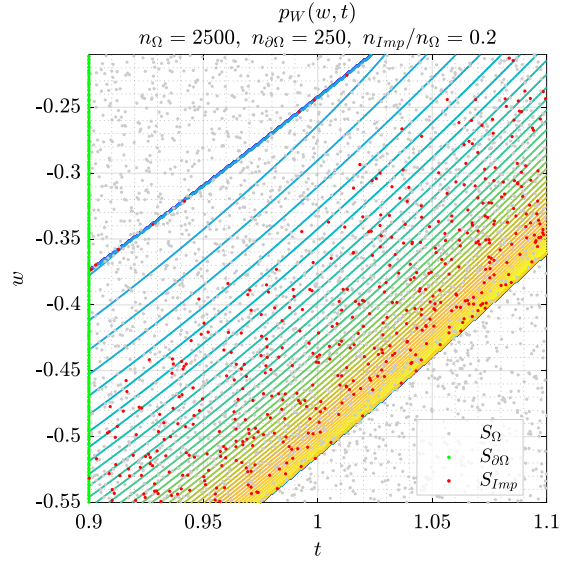

Figure 14. EB-beam forced vibration spatiotemporal collocation points: LHS-generated, IC, and importance-sampled points

Once again, the LSTM-LBFGS network shows the best accuracy as can also be seen in Table 3. Figure 15 illustrates a comparison between the estimated instantaneous PDF at $t = 1$. The MLP-Adam and the MLP-Swish-LBFGS networks show oscillatory behavior which can be indicative of overfitting for these networks.

Table 3 Case-Study (III) NN Depth = 4, NN Width = 20, Collocation Points = 2500, Epochs = 50000

| Network | Sampling Fraction = 0.0 | | Sampling Fraction = 0.2 | | Sampling Fraction = 0.5 | |
|---|---|---|---|---|---|---|
| | Loss | N.E.T. | Loss | N.E.T. | Loss | N.E.T. |
| MLP-LBFGS | 0.055811 | 0.804099 | 0.008476 | 1* | 0.076180 | 1.377885 |
| LSTM-LBFGS | 0.012828 | 1.117091 | 0.042502 | 1.391124 | 0.001768 | 1.724116 |
| MLP-Adam | 0.599014 | 0.647373 | 0.432074 | 0.844874 | 0.227557 | 1.039 |
| LSTM-Adam | 0.005974 | 0.957603 | 0.001952 | 1.302596 | 0.000023 | 1.525549 |
| MLP-Swish-LBFGS | 0.153767 | 0.933784 | 0.180830 | 1.118949 | 0.012612 | 1.497481 |
| (*) t=51636 s | | | | | | |

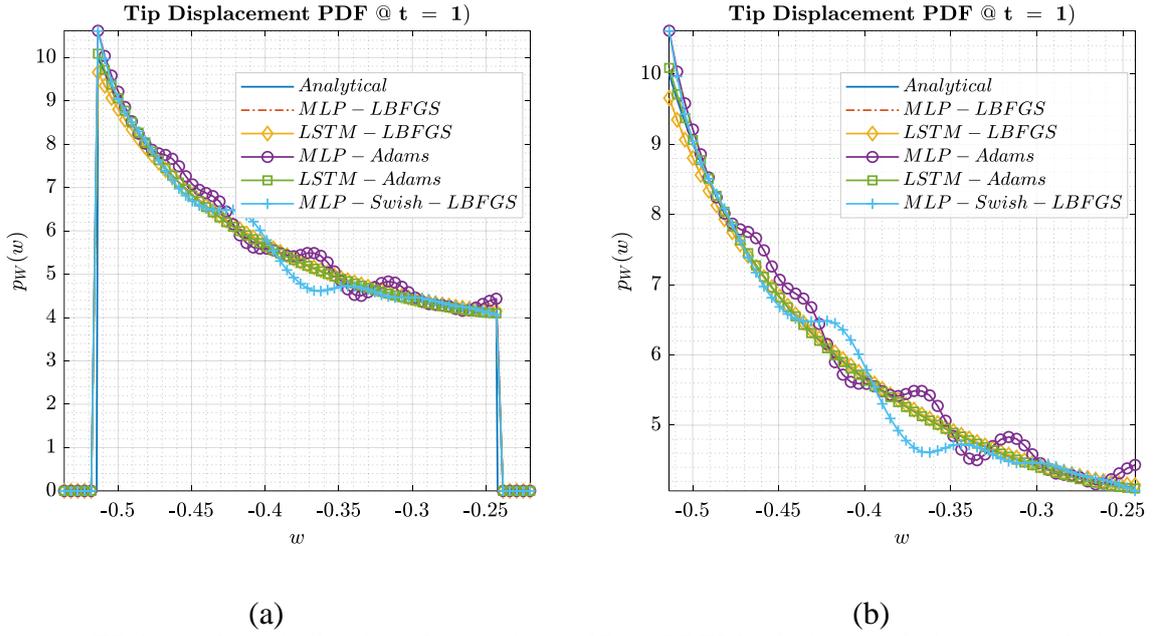

(a)            (b)

Figure 15. EB-beam forced vibration: Comparison of DeepPDEM solutions with the exact solution at t = 1. (a) Overall behavior of the PDF, (b) PDF behavior with scaled up vertical axis limits

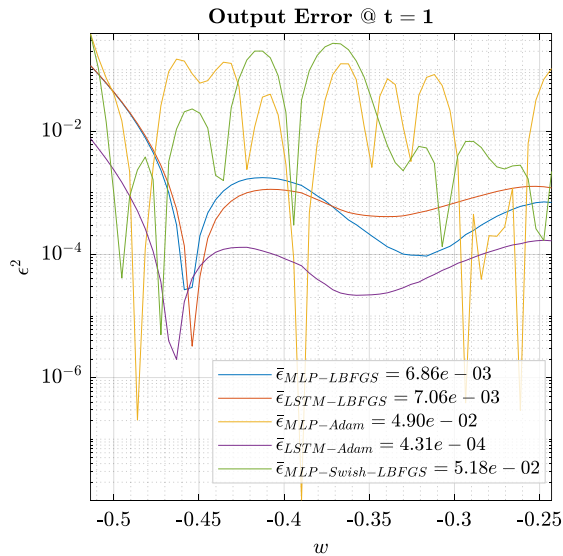

Figure 16. EB-beam forced vibration: Comparison of DeepPDEM estimation error with respect to the exact solution at t = 1

- *Sensitivity Analysis*

In this test case, the sample mean is $\mu = 0.0052$, and the sample variance is $S^2 = 2.2e-5$, and sample skewness is $G_1 = 1.9131$. Hence, as the result shown in Figure 17 suggests, there is no strong sensitivity to the initialized weights.

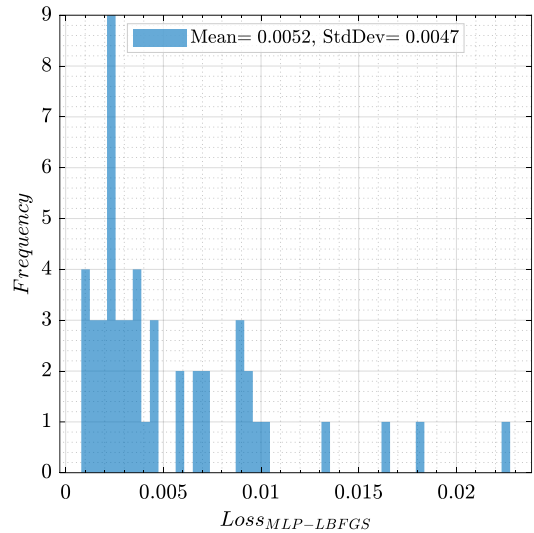

Figure 17. EB-beam forced vibration: variation of DeepPDEM loss with respect to Glorot normal initializer utilizing different random seeds in 50 Monte-Carlo experiments

Figure 18 presents the results found for the learning rate study. As indicated in this Figure 18, choosing a learning rate in the interval $[5e-4, 5e-3]$ provides acceptable results.

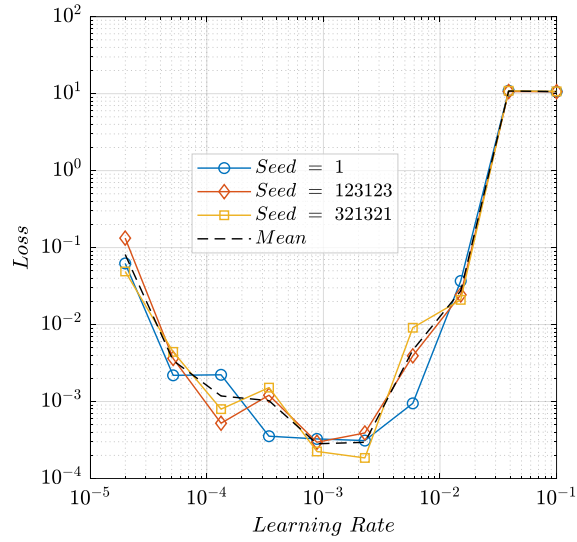

Figure 18. EB-beam forced vibration: variation of DeepPDEM loss with respect to different rectified adaptive learning rates considering three different seeds

## 4.3 Results and Discussion

The current investigation has demonstrated that the proposed DeepPDEM can effectively be utilized as a surrogate model of GDEE. The results indicate that using LSTM architecture within the DeepPDEM framework can improve the accuracy of the results. However, the latter comes at an increasing cost of the training. On the other hand, the results also indicate that using a better-suited activation function can compensate for the accuracy of the MLP architecture. Furthermore, it is evident that in some cases using the L-BFGS-B optimizer increases the chance of premature convergence. Even though DNNs can easily overfit, they also tend to generalize relatively well [49] which requires further studies.

Through the current investigation, it was observed that considering long-time integration, training of the network becomes exceedingly expensive. This is due to the fact that as the integration time increases the GDEE spatiotemporal domain grows rapidly. The DNNs are parallel in na-ture [50] which can be leveraged to achieve long-time integrations. Currently, physics-constrained networks are slower than other numerical solvers such as finite elements. However using offline training, this limitation is highly compensated [51]. Also, choosing an effective architecture for the DNN is still an open problem and is usually addressed empirically. Meta-learning is an emerging research topic that may also resolve this issue [52].

# 5 Concluding Remarks

The current study proposes DeepPDEM, a physics-informed deep neural network methodology that solves and serves as a surrogate model for General Density Evolution Equation (GDEE) for stochastic structural systems. By using automatic differentiation, the differential operators in GDEE are handled to achieve a mesh-free solver for the probability density evolution problem. Assuming that the gradients estimated by the DeepPDEM tend to the solution gradient, they are utilized to enhance the sampling and improve the training efficiency. Moreover, the performance of two different DNN architectures is investigated within the DeepPDEM framework. To test, verify and benchmark the proposed method, DeepPDEM is utilized to solve the density evolution of three relevant test cases whose analytical exact solutions are derived for comparison purposes. The comparisons are indicative of excellent agreement between the DeepPDEM surrogate and the analytical solutions.

Finally, DeepPDEM can be used for uncertainty quantification, design optimization, and control policy design of stochastic structures. On the other hand, DNNs usually demand a large quantity of labeled data to be trained that may not be available or be expensive to compile. In this respect, the proposed DeepPDEM can be a novel surrogate for the GDEE that does not need any labeled data for training purposes. In essence, the proposed DeepPDEM is trained solely based on minimizing the GDEE in the residual form without using any labeled data. DeepPDEM can accurately predict the probability density evolution with various ICs and BCs, to propagate uncertainties more efficiently. The results potentially demonstrate the good merits of DeepPDEM and pave the way for its further development as a surrogate model in complex applications requiring real-time probability density evolution.

## Appendix. A – Free Vibration of an Uncertain Single Degree of Freedom System

For the free vibration of an uncertain 1DOF system one can propose the following system:

$$\ddot{x} + \omega^2 x = 0 \; ; \; \dot{x}(0) = \dot{x}_0 \; ; \; x(0) = x_0 \; ; \; \omega \sim U(\omega_1, \omega_2) \tag{25}$$

The analytical solution for this second-order homogeneous ODE is given as:

$$\begin{cases} x(t) = \dfrac{\dot{x}_0}{\omega} \sin(\omega t) + x_0 \cos(\omega t) \\ \dot{x}(t) = \dot{x}_0 \cos(\omega t) - x_0 \omega \sin(\omega t) \end{cases} \tag{26}$$

Here we only consider the PDF evolution of the position. The evolution partial differential equation can be written as:

$$\frac{\partial p_{X,\Omega}(x, \omega, t)}{\partial t} + (\dot{x}_0 \cos(\omega t) - x_0 \omega \sin(\omega t)) \cdot \frac{\partial p_{X,\Omega}(x, \omega, t)}{\partial x} = 0 \tag{27}$$

$$\text{I. C.:} \quad p_{X,\Omega}(x, \omega, t_0) = \delta(x - x_0) p_\Omega(\omega) \tag{28}$$

Since the PDF of the natural frequency is uniformly distributed over $[\omega_1, \omega_2]$:

$$p_\Omega(\omega) = \frac{1}{\omega_2 - \omega_1}(\mathbb{H}(\omega - \omega_1) - \mathbb{H}(\omega - \omega_2)) = \begin{cases} \dfrac{1}{\omega_2 - \omega_1} & \omega \in [\omega_1, \omega_2] \\ 0 & \omega \notin [\omega_1, \omega_2] \end{cases} \tag{29}$$

Thus, the GDEE would be:

$$\begin{cases} \dfrac{\partial p_{X,\Omega}(x,\omega,t)}{\partial t} + (\dot{x}_0 \cos(\omega t) - x_0 \omega \sin(\omega t)) \cdot \dfrac{\partial p_{X,\Omega}(x,\omega,t)}{\partial x} = 0 \\ \text{I.C.:} \quad p_{X,\Omega}(x,\omega,t_0 = 0) = \dfrac{(\mathbb{H}(\omega - \omega_1) - \mathbb{H}(\omega - \omega_2))}{(\omega_2 - \omega_1)} \delta(x - x_0) \end{cases} \quad (30)$$

This is a linear hyperbolic PDE that can be solved using the method of characteristics.

$$p = f\left(x - \frac{\dot{x}_0}{\omega} \sin(\omega t) - x_0 \cos(\omega t)\right) \quad (31)$$

where "f" is an arbitrary function. Note that "$p$" is constant along the characteristic curves.

From the initial condition:

$$p_{X,\Omega}(x,\omega,t_0 = 0) = \frac{(\mathbb{H}(\omega - \omega_1) - \mathbb{H}(\omega - \omega_2))}{(\omega_2 - \omega_1)} \delta(x - x_0) \quad (32)$$

Therefore, the solution can be suggested as:

$$p_{X,\Omega}(x,\omega,t) = \frac{(\mathbb{H}(\omega - \omega_1) - \mathbb{H}(\omega - \omega_2))}{(\omega_2 - \omega_1)} \delta\left(x - \frac{\dot{x}_0}{\omega} \sin(\omega t) - x_0 \cos(\omega t)\right) \quad (33)$$

$$p_X(x,t) = \frac{1}{(\omega_2 - \omega_1)} \int_{\omega_1}^{\omega_2} \delta\left(x - \frac{\dot{x}_0}{\omega} \sin(\omega t) - x_0 \cos(\omega t)\right) d\omega \quad (34)$$

To integrate (34) we first need to determine the roots of the function inside the Dirac delta:

$$\delta(f(\omega)) = \sum_{\kappa} \frac{1}{|f'(\omega_\kappa)|} \delta(\omega - \omega_\kappa) \quad : \quad f(\omega_\kappa) = 0 \quad (35)$$

$$f'(\omega) = \frac{\partial f(\omega)}{\partial \omega} = x_0 t \sin(\omega t) - \frac{\dot{x}_0 t}{\omega} \cos(\omega t) + \frac{\dot{x}_0 t}{\omega^2} \sin(\omega t) \quad (36)$$

For the brevity of notation let's assume:

$$C_\kappa = \frac{1}{(\omega_2 - \omega_1) \cdot \left| x_0 t \sin(\omega_\kappa t) - \frac{\dot{x}_0 t}{\omega_\kappa} \cos(\omega_\kappa t) + \frac{\dot{x}_0 t}{\omega_\kappa^2} \sin(\omega_\kappa t) \right|} \quad (37)$$

$$\delta\bigl(f(\omega)\bigr) = (\omega_2 - \omega_1)\sum_{\kappa} C_\kappa \delta(\omega - \omega_\kappa) \quad : \quad f(\omega_\kappa) = 0 \tag{38}$$

Therefore:

$$p_X(x,t) = \int \sum_{\kappa} C_\kappa \cdot \delta(\omega - \omega_\kappa) \cdot (\mathbb{H}(\omega - \omega_1) - \mathbb{H}(\omega - \omega_2))d\omega \tag{39}$$

Since the summations and integration are interchangeable for integrands greater than or equal to zero (Tonelli's theorem):

$$p_X(x,t) = \sum_{\kappa} C_\kappa \cdot \int \delta(\omega - \omega_\kappa) \cdot (\mathbb{H}(\omega - \omega_1) - \mathbb{H}(\omega - \omega_2))d\omega \tag{40}$$

$$p_X(x,t) = \sum_{\kappa} C_\kappa \cdot \bigl(\mathbb{H}(\omega_\kappa - \omega_1) - \mathbb{H}(\omega_\kappa - \omega_2)\bigr) \tag{41}$$

For the case where the initial velocity is equal to zero, we have:

$$p_{X,\Omega}(x,\omega,t) = \frac{(\mathbb{H}(\omega - \omega_1) - \mathbb{H}(\omega - \omega_2))}{(\omega_2 - \omega_1)}\delta(x - x_0\cos(\omega t)) \tag{42}$$

$$p_{X,\Omega}(x,\omega,t) = \frac{(\mathbb{H}(\omega - \omega_1) - \mathbb{H}(\omega - \omega_2))}{(\omega_2 - \omega_1)}$$

$$\cdot \sum_{\kappa} \frac{1}{\left|x_0 t \sin\left(\cos^{-1}\frac{x}{x_0} + 2\kappa\pi\right)\right|} \delta\left(\omega - \frac{1}{t}\left(\cos^{-1}\frac{x}{x_0} + 2\kappa\pi\right)\right) \tag{43}$$

Considering a case where $x_0 = 0.1$, $\omega_1 = \frac{\pi}{4}$, $\omega_2 = \frac{3\pi}{4}$, and $t \in [0.9, 1.1]$ leaves us with:

$$p_X(x,t) = \frac{\left(\mathbb{H}\left(\frac{1}{t}\cos^{-1}10x - \frac{\pi}{4}\right) - \mathbb{H}(\frac{1}{t}\cos^{-1}10x - \frac{3\pi}{4})\right)}{\frac{\pi}{2}|0.1t\sin(\cos^{-1}10x)|} \tag{44}$$

# Appendix. B – Free Vibrations of a Cantilever Euler-Bernoulli Beam

In this case, the uncertainty of the beam flexural rigidity is assumed to be uniformly distributed.

Thus, the equations of motion for the free vibration of a beam can be written as:

$$\begin{cases} \dfrac{\partial^2 w}{\partial t^2} + \alpha^2 \dfrac{\partial^4 w}{\partial x^4} = 0 \\ w(0,t) = w_x(0,t) = 0 \quad ; \quad \alpha \sim U(\alpha_1, \alpha_2) \\ w_{xx}(l,t) = w_{xxx}(l,t) = 0 \\ \qquad w(x,0) = \xi(x) \\ \qquad w_t(x,0) = v(x) \end{cases} \quad (45)$$

Therefore, for the GDEE we have:

$$\dot{W} = \alpha \mu_n^2 \sum_{n=1}^{\infty} \left(-a_n \sin(\alpha \mu_n^2 t) + b_n \cos(\alpha \mu_n^2 t)\right)$$

$$\cdot \left( (\cos \mu_n x - \cosh \mu_n x) - \frac{\sin \mu_n l + \sinh \mu_n l}{\cos \mu_n l + \cosh \mu_n l} \right. \quad (46)$$

$$\left. \cdot (\sin \mu_n x - \sinh \mu_n x) \right) 0$$

$$\begin{cases} \dfrac{\partial \wp_{W,A}(w, \alpha, t)}{\partial t} + \dot{W} \cdot \dfrac{\partial \wp_{W,A}(w, \alpha, t)}{\partial w} = 0 \\ \text{I.C.:} \quad \wp_{W,A}(w, \alpha, t_0 = 0) = \dfrac{\left(\mathbb{H}(\alpha - \alpha_1) - \mathbb{H}(\alpha - \alpha_2)\right)}{(\alpha_2 - \alpha_1)} \delta(w - w_0) \end{cases} \quad (47)$$

$$p_{W,A}(w, \alpha, t) = \frac{\left(\mathbb{H}(\alpha - \alpha_1) - \mathbb{H}(\alpha - \alpha_2)\right)}{(\alpha_2 - \alpha_1)} \delta\Bigg(w$$
$$- \sum_{n=1}^{\infty} (a_n \cos(\alpha\mu_n^2 t) + b_n \sin(\alpha\mu_n^2 t))$$
$$\cdot \Bigg((\cos\mu_n x - \cosh\mu_n x) - \frac{\sin\mu_n l + \sinh\mu_n l}{\cos\mu_n l + \cosh\mu_n l}$$
$$\cdot (\sin\mu_n x - \sinh\mu_n x)\Bigg)\Bigg) \tag{48}$$

We may simplify the problem by considering the following initial conditions:

$$\xi(x) = (\cos\mu_1 x - \cosh\mu_1 x) - \frac{\sin\mu_1 l + \sinh\mu_1 l}{\cos\mu_1 l + \cosh\mu_1 l} \cdot (\sin\mu_1 x - \sinh\mu_1 x) \tag{49}$$

$$v(x) = \alpha\mu_1^2 \cdot \Bigg((\cos\mu_1 x - \cosh\mu_1 x) - \frac{\sin\mu_1 l + \sinh\mu_1 l}{\cos\mu_1 l + \cosh\mu_1 l} \cdot (\sin\mu_1 x - \sinh\mu_1 x)\Bigg) \tag{50}$$

Therefore, $a_1 = b_1 = 1$ while $a_i = b_i = 0$, $2 \leq i$. We may also keep track of only the tip deflections.

$$p_{W,A}(w, \alpha, t) = \frac{\left(\mathbb{H}(\alpha - \alpha_1) - \mathbb{H}(\alpha - \alpha_2)\right)}{(\alpha_2 - \alpha_1)} \delta\Bigg(w - 2\frac{\cos^2\mu_1 l - 1}{\cos\mu_1 l + \cosh\mu_1 l}$$
$$\cdot (\cos(\alpha\mu_1^2 t) + \sin(\alpha\mu_1^2 t))\Bigg) \tag{51}$$

We can rewrite $\cos(\alpha\mu_1^2 t) + \sin(\alpha\mu_1^2 t)$ as $\sqrt{2}\sin\left(\alpha\mu_1^2 t + \frac{\pi}{4}\right)$, that gives:

$$p_{W,A}(w, \alpha, t) = \frac{\left(\mathbb{H}(\alpha - \alpha_1) - \mathbb{H}(\alpha - \alpha_2)\right)}{(\alpha_2 - \alpha_1)}$$
$$\cdot \sum_\kappa \frac{\delta(\alpha - \alpha_\kappa)}{\left|2\sqrt{2}\mu_1^2 t \frac{\cos^2\mu_1 l - 1}{\cos\mu_1 l + \cosh\mu_1 l} \cdot \cos\left(\alpha_\kappa \mu_1^2 t + \frac{\pi}{4}\right)\right|} \tag{52}$$

Let's consider a case where $l = 1$, $0.9 \leq t \leq 1.1$, and $\alpha \sim U\left(\frac{\pi}{4}, \frac{3\pi}{4}\right)$. For these numerical values, we will have the density evolution solution as:

$$p_W(w, t) = \sum_\kappa \frac{\mathbb{H}\left(\alpha_\kappa - \frac{\pi}{4}\right) - \mathbb{H}\left(\alpha_\kappa - \frac{3\pi}{4}\right)}{\frac{\pi}{2}\left|0.8475t \cdot \cos\left(\alpha_\kappa \mu_1^2 t + \frac{\pi}{4}\right)\right|} \tag{53}$$

## Appendix. C – Free Vibration of a Cantilever Euler-Bernoulli Beam

The motion is governed by the following PDE:

$$\frac{\partial^2 w}{\partial t^2} + \alpha^2 \frac{\partial^4 w}{\partial x^4} = q(x, t) \tag{54}$$

Assuming a modal approach for the solution:

$$w(x, t) = \sum_{i=1}^{\infty} \phi_i(x) \eta_i(t) \tag{55}$$

where $\phi_i$ are the normal modes found by solving the equation using the four boundary conditions and $\eta_i$ are the generalized coordinates or modal participation coefficients. Therefore, we will have:

$$w(x, t) = \sum_{n=1}^{\infty} \phi_n(x) \left( a_n \sin \omega_n t + b_n \cos \omega_n t + \frac{1}{\omega_n} \int_0^t \int_0^l \phi_n(x) q(x, \tau) \sin(\omega_n \cdot (t - \tau)) \, dx d\tau \right) \tag{56}$$

Let's assume that the beam is subjected to a distributed force similar to its $m^{th}$ normal mode. Considering uncertainty with uniform distribution for the excitation phase shift, the EOM can be written as:

$$\begin{cases} \frac{\partial^2 w}{\partial t^2} + \alpha^2 \frac{\partial^4 w}{\partial x^4} = q_0 \sin(\Omega t + \psi) \phi_m(x) \\ \quad w(0, t) = w_x(0, t) = 0 \qquad ; \quad \psi \sim U(\psi_1, \psi_2) \\ \quad w_{xx}(l, t) = w_{xxx}(l, t) = 0 \\ \quad w(x, 0) = \xi(x) \\ \quad w_t(x, 0) = v(x) \end{cases} \tag{57}$$

$$\frac{1}{\omega_n}\int_0^t\int_0^l \phi_n(x)q_0 \sin(\Omega t + \psi)\, \phi_m(x)\, \sin(\omega_n \cdot (t - \tau))\, dx dt$$

$$= \frac{q_0}{\omega_m}\int_0^t \sin(\Omega\tau + \psi) \sin(\omega_m \cdot (t - \tau))\, d\tau$$

$$= \frac{q_0}{\Omega^2 - \omega_m^2}\left(\left(\frac{\Omega}{\omega_m}\sin(\omega_m t) - \sin(\Omega t)\right)\cos(\psi)\right.$$

$$\left. + (\cos(\omega_m t) - \cos(\Omega t))\sin(\psi)\right) \quad (58)$$

where:

$$\phi_n(x) = (\cos \mu_n x - \cosh \mu_n x) - \frac{\sin \mu_n l + \sinh \mu_n l}{\cos \mu_n l + \cosh \mu_n l} \cdot (\sin \mu_n x - \sinh \mu_n x) \quad (59)$$

Therefore, the modal participation coefficients can be expressed as:

$$\eta_n(t) = a_n \cdot \sin \omega_n t + b_n \cos \omega_n t$$

$$+ \frac{q_0}{\Omega^2 - \omega_m^2}\left(\left(\frac{\Omega}{\omega_m}\sin(\omega_m t) - \sin(\Omega t)\right)\cos(\psi)\right.$$

$$\left. + (\cos(\omega_m t) - \cos(\Omega t))\sin(\psi)\right) \quad (60)$$

We may simplify the problem by considering the following initial conditions:

$$\xi(x) = (\cos \mu_j x - \cosh \mu_j x) - \frac{\sin \mu_j l + \sinh \mu_j l}{\cos \mu_j l + \cosh \mu_j l} \cdot (\sin \mu_j x - \sinh \mu_j x) = \phi_j(x) \quad (61)$$

$$v(x) = \alpha\mu_j^2 \cdot \left((\cos \mu_j x - \cosh \mu_j x) - \frac{\sin \mu_j l + \sinh \mu_j l}{\cos \mu_j l + \cosh \mu_j l} \cdot (\sin \mu_j x - \sinh \mu_j x)\right)$$

$$= \omega_j^2 \cdot \phi_j(x) \quad (62)$$

This assumption forces that, $a_j = b_j = 1$ while $a_i = b_i = 0$, $2 \leq i$. We may also keep track of only the tip deflections. Therefore, the response of the beam can be expressed as:

$$w(x,t) = \left(\sin \omega_j t + \cos \omega_j t\right) \cdot \phi_j(x)$$

$$+ \frac{q_0}{\Omega^2 - \omega_m^2} \left(\left(\frac{\Omega}{\omega_m}\sin(\omega_m t) - \sin(\Omega t)\right)\cos(\psi)\right.$$

$$\left. + \left(\cos(\omega_m t) - \cos(\Omega t)\right)\sin(\psi)\right) \cdot \phi_m(x) \quad (63)$$

We may now consider the GDEE equation defining $\dot{W}$ as:

$$\dot{W} = \omega_j\left(\cos \omega_j t - \sin \omega_j t\right) \cdot \phi_j(x)$$

$$+ \frac{q_0}{\Omega^2 - \omega_m^2} \left(\left(\frac{\Omega}{\omega_m}\sin(\omega_m t) - \sin(\Omega t)\right)\cos(\psi)\right.$$

$$\left. + \left(\cos(\omega_m t) - \cos(\Omega t)\right)\sin(\psi)\right) \cdot \phi_m(x) \quad (64)$$

$$\begin{cases} \dfrac{\partial p_{W,\Psi}(w,\psi,t)}{\partial t} + \dot{W} \cdot \dfrac{\partial p_{W,\Psi}(w,\psi,t)}{\partial w} = 0 \\ \text{I.C.:} \quad p_{W,\Psi}(w,\psi,t_0 = 0) = \dfrac{\mathbb{H}(\psi - \psi_1) - \mathbb{H}(\psi - \psi_2)}{\psi_2 - \psi_1}\delta(w - w_0) \end{cases} \quad (65)$$

$$p_{W,\Psi}(w,\psi,t) = \frac{\mathbb{H}(\psi - \psi_1) - \mathbb{H}(\psi - \psi_2)}{\psi_2 - \psi_1}\delta\left(w - \left(\sin \omega_j t + \cos \omega_j t\right)\right.$$

$$\cdot \phi_j(x)$$

$$- \frac{q_0}{\Omega^2 - \omega_m^2}\left(\left(\frac{\Omega}{\omega_m}\sin(\omega_m t) - \sin(\Omega t)\right)\cos(\psi)\right.$$

$$\left.\left. + \left(\cos(\omega_m t) - \cos(\Omega t)\right)\sin(\psi)\right) \cdot \phi_m(x)\right) \quad (66)$$

Defining $c_i$ as:

$$c_1 = \frac{\phi_m}{\omega_m^2}\left(\cos(\omega_m t) - \cos(\Omega t)\right) \quad (67)$$

$$c_2 = \frac{\phi_m}{\omega_m^3}(4\sin(\omega_m t) - \omega_m \sin(\Omega t)) \tag{68}$$

$$c_3 = \left(w - \phi_j(\sin(\omega_j t) + \cos(\omega_j t))\right) \tag{69}$$

The Dirac delta input function can be rewritten as:

$$f(\psi) = c_1 \sin\psi + c_2 \cos\psi + c_3 = 0 \tag{70}$$

$$f'(\psi) = \frac{\partial f(\psi)}{\partial \psi} = c_1 \cos\psi - c_2 \sin\psi \tag{71}$$

The solution to this equation can be found as:

$$\psi_{\kappa,1} = \tan^{-1}\left(\frac{-c_1^2 c_3 - c_2\sqrt{c_1^2(c_1^2 + c_2^2 - c_3^2)}}{\left(-c_2 c_3 + \sqrt{c_1^2(c_1^2 + c_2^2 - c_3^2)}\right)c_1}\right) \tag{72}$$

$$\psi_{\kappa,2} = \tan^{-1}\left(\frac{c_1^2 c_3 - c_2\sqrt{c_1^2(c_1^2 + c_2^2 - c_3^2)}}{\left(c_2 c_3 + \sqrt{c_1^2(c_1^2 + c_2^2 - c_3^2)}\right)c_1}\right) \tag{73}$$

Assume $j = m = 1$, $\alpha = 1$, $\Omega = 2$, $q_0 = 1$, and $l = 1$. Once again, we only keep track of the beam motion at the tip. Hence, the density evolution solution will be:

$$p_{W,\Psi}(w, \psi, t) = \frac{\mathbb{H}(\psi - \psi_1) - \mathbb{H}(\psi - \psi_2)}{\psi_2 - \psi_1} \cdot \sum_\kappa \frac{\delta(\psi - \psi_\kappa)}{|f'(\psi_\kappa)|} \tag{74}$$

$$p_W(w, t) = \sum_\kappa \frac{(\mathbb{H}(\psi_\kappa - \psi_1) - \mathbb{H}(\psi_\kappa - \psi_2))/(\psi_2 - \psi_1)}{|f'(\psi_\kappa)|} \tag{75}$$